\documentclass[10pt,twocolumn,letterpaper]{article}

\usepackage{multirow}
\usepackage[pagenumbers]{cvpr}

\definecolor{cvprblue}{rgb}{0.21,0.49,0.74}
\usepackage[pagebackref,breaklinks,colorlinks,allcolors=cvprblue]{hyperref}

\def\paperID{*****}
\def\confName{CVPR}
\def\confYear{2026}

\title{Treatment-Conditioned Diffusion for Forecasting Neurodegenerative Disease Progression}

\author{
Danylo Boiko\\
Innoloft Inc.\\
701 Brazos Street, Austin, TX 78701, USA\\
{\tt\small d.boiko@innoloft.com}
\and
Viktoriia Mishkurova\\
Bogomolets National Medical University\\
Beresteiskyi Avenue, 34, Kyiv, 03057, Ukraine\\
{\tt\small michkourovaviktoire@gmail.com}
}

\begin{document}
\maketitle
\begin{abstract}
Forecasting the progression of neurodegenerative diseases, such as Parkinson's disease, is essential for effective long-term planning and personalized therapeutic intervention. Existing systems typically produce scalar clinical scores that ignore the rich structure of longitudinal neuroimaging, while traditional generative approaches suffer from a loss of anatomical details and blurring subtle progression patterns. To address this, we introduce a novel treatment-conditioned diffusion framework that predicts high-fidelity future brain states by conditioning the generative process on patients' screening DaTscan images and levodopa equivalent daily dose over one year. The pipeline uses a Transformer-based encoder to represent non-linear, time-dependent pharmacological dynamics and optimizes generation through a multi-weight region-of-interest mask that focuses on biologically critical areas. Experimental evaluation shows that our framework maintains sharp anatomical boundaries and significantly improves clinical fidelity relative to the baseline, achieving 14.0\% lower MSE, 7.2\% lower MAE, and 4.9\% higher SSIM.
\end{abstract}    
\section{Introduction}
\label{sec:intro}

Modeling the progression of neurodegenerative diseases, including Parkinson's, is a key factor in the personalization and long-term planning of therapeutic strategies. Parkinson's disease (PD) is one of the leading neurodegenerative diseases worldwide and the fastest-growing among them~\cite{Dorsey2018}. This condition is associated with a wide range of motor and non-motor symptoms that significantly reduce quality of life. Early diagnosis and therapeutic intervention are correlated with slower progression and delayed onset of certain symptoms.

Classical predictive techniques ranging from multilayer perceptrons (MLPs) and recurrent neural networks (RNNs) to more recent graph-based or ordinary differential equation (ODE) approaches typically output scalar clinical scores (\eg, UPDRS, MMSE, MoCA) while ignoring the rich structure of longitudinal neuroimaging (\eg, MRI, CT, PET, SPECT). Generative approaches such as generative adversarial networks (GANs) and variational autoencoders (VAEs) are useful for synthesizing brain images~\cite{Tudosiu2024}. However, they remain plagued by loss of anatomical details and fuzziness of subtle progression patterns.

In this work, we introduce a novel treatment-conditioned diffusion paradigm. Building on the success of conditional diffusion models (CDMs), which have achieved state-of-the-art results in high-fidelity medical image synthesis~\cite{zhan2025}, we propose leveraging their generative power to forecast PD progression by conditioning the diffusion process on patients' data. In particular, the model learns the conditional distribution based on DaTscan images, a type of SPECT imaging used to evaluate dopamine transporters, and the monthly levodopa equivalent daily dose (LEDD) over the course of a year. To implement this framework, we designed a lightweight 2D U-Net~\cite{si2023} that predicts future DaTscan images by channel-wise concatenation of noise and screening DaTscan images, while incorporating LEDD embeddings via a projection layer.

Key contributions are highlighted below:
\begin{itemize}
    \item Novel diffusion framework that forecasts brain states conditioned on screening DaTscan images and treatments.
    \item Contrastive Transformer-based autoencoder for noisy, time-dependent medication histories.
    \item Multi-weight ROI optimization pipeline that forces the model to prioritize biologically critical regions.
    \item Improvement of clinical fidelity, achieving 14.0\% lower MSE, 7.2\% lower MAE, and 4.9\% higher SSIM across 14 slices compared to the baseline.
\end{itemize}

\section{Related work}
\label{sec:related_work}

Over the last decades, the mathematical and computational modeling of neurodegenerative disease progression has shifted from linear clinical observations toward complex longitudinal frameworks that account for multi-modal biological decay. The central element of this evolution is the merging of diverse data streams ranging from longitudinal imaging and genetic markers to cerebrospinal fluid (CSF) proteomics~\cite{Lian2024,Severson2021}. These models assume that disease-specific biomarkers follow predictable paths that can be captured through latent-variable modeling or state-space representations~\cite{Tabashum2024,Oxtoby2023}. By synchronizing patients at different clinical stages, we can identify early-stage physiological changes that precede a clear symptomatic decline, providing a critical window for medical treatment~\cite{Wang2025,Oxtoby2023}.

Although statistical methods provide interpretability, recent advances in machine learning (ML) and deep learning (DL) offer a notable capacity to handle the high-dimensional, non-Gaussian nature of clinical datasets~\cite{Severson2021}. For example, graph neural networks (GNNs) and neural ODE models are increasingly utilized to predict individual-level brain atrophy or cognitive scores by capturing long-range dependencies in longitudinal data~\cite{Lian2023}. These approaches excel in identifying subtle, non-linear relationships between risk factors, such as the combined effect of genetic mutations (\eg, in GBA or LRRK2) and motor symptom progression~\cite{Lian2024,Wang2025}. However, a persistent limitation of these models is their lack of a biological basis, unlike mechanistic biophysical models~\cite{Tabashum2024}. Consequently, there is a growing convergence toward hybrid systems that constrain ML outputs with known physiological priors to ensure clinical relevance~\cite{Lian2023,He2025}. 

GANs and VAEs demonstrated excellent capabilities for cross-modality synthesis and the augmentation of small datasets, enabling the creation of high-accuracy anatomical transitions~\cite{Friedrich2025}. This predictive power is further refined by denoising diffusion probabilistic models (DDPMs), which provide a mathematically solid framework for generating diverse synthetic brain images while maintaining the structural integrity of pathological markers~\cite{Khader2023}. Despite the strength of these models in capturing non-linear morphological changes, a major area of convergence remains the requirement to balance computational complexity with clinical interpretability, ensuring that synthetic outputs align with the biological priors of neurodegeneration.

Nowadays, applications of diffusion models have evolved from image synthesis to forecasting the progression of neurodegenerative diseases, offering a robust probabilistic framework to predict future brain states~\cite{Zhao2025}. While a substantial number of systems used this technique for cross-sectional normative modeling~\cite{Whitbread2026} and unsupervised anomaly detection~\cite{Aguila2025,Behrendt2025}, the methodological focus is rapidly expanding toward longitudinal predictive modeling. To enforce precise anatomical constraints throughout image generation, novel pipelines use segmentation-guided sampling and mask-aware training, ensuring high fidelity while allowing localized flexibility~\cite{Konz2024}.

By framing disease progression as a conditional generation task, CDMs can synthesize highly realistic future neuroimaging representations that reflect anticipated pathological changes, overcoming the weaknesses of deterministic predictive algorithms~\cite{Dao2024,Xiao2024}. The main methodological assumption here is that patients' data provide a sufficient predictive signal to dictate localized, time-dependent pathological deformations. Consequently, while generative models show significant strengths in forecasting subject-specific trajectories with narrow confidence intervals, their predictive accuracy can degrade while forecasting highly non-linear neurodegeneration or unobserved variants~\cite{Xiao2024}.
\section{Data and preprocessing}
\label{sec:data}

Data used in this paper were obtained on September 28, 2025, from the Parkinson’s progression markers initiative (PPMI) database~\cite{Marek2018}, an ongoing longitudinal observational study launched in 2010 by the Michael J. Fox Foundation for Parkinson's research. PPMI aims to identify biomarkers for the onset and progression of PD by collecting multimodal data from diverse cohorts.

Levodopa is the most effective and widely prescribed treatment for PD. It is a precursor to dopamine that can pass through the blood-brain barrier, thus restoring dopamine levels in the striatum, reducing motor symptoms, and improving patients' quality of life~\cite{fneur2018}.  We selected LEDD as a condition because it reflects the cumulative dopaminergic effect on the brain throughout the day, rather than temporary peaks from single doses. 

Using DaTscan images in combination with LEDD allows to directly correlate structural and functional changes in the basal ganglia with the patient's clinical condition. Moreover, by focusing on slices 34 to 47, which cover the caudate nucleus and putamen, we target the most affected areas while reducing noise from less informative regions, increasing the accuracy and statistical significance.

After applying basic filters, we identified 803 subjects who participated in the PD study and 9,147 medication records from the entire database. Filtering for patients who remained on medication from screening through month 12 resulted in a final cohort of 212 participants. Each of them has 14 DaTscan images of size 128$\times$128 pixels (slices 34 to 47) for both stages and a 12-dimensional LEDD vector. Accordingly, a total of 2,968 pairs are available for the training, validation, and test subsets.

\begin{figure*}
  \centering
  \includegraphics[width=\linewidth]{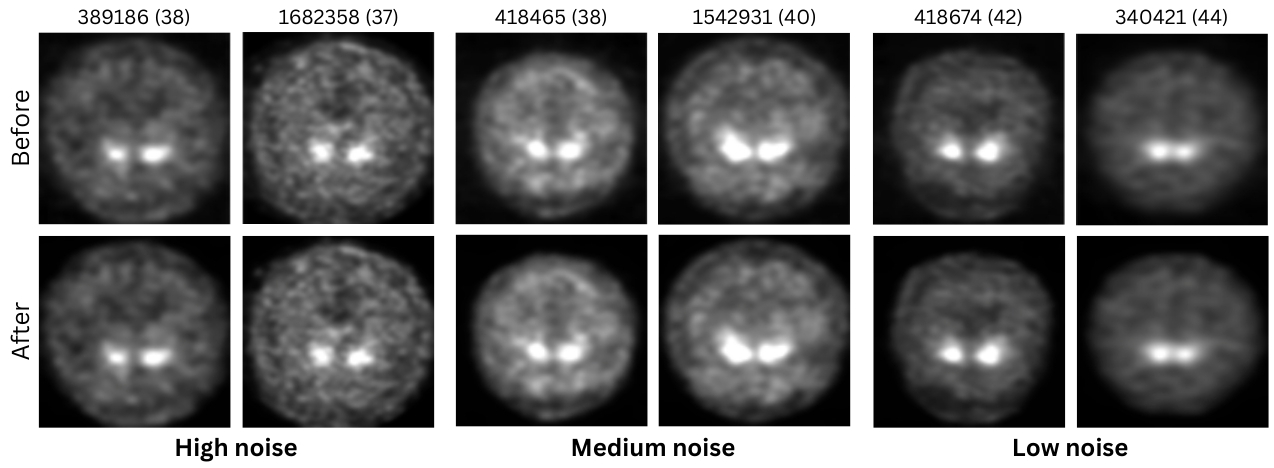}
  \caption{Comparison of DaTscan images before and after preprocessing at different noise levels (high, medium, low). The number above each DaTscan is the image ID, while the number in parentheses is the slice number.}
  \label{fig:preprocessing}
\end{figure*}

To improve image quality and mitigate the background noise and motion artifacts inherent to DaTscan isotope registration~\cite{Rahmim2013}, we employ a dynamic soft-masking preprocessing mechanism (see \cref{fig:preprocessing}). The final composite tensor, $I_{out}$, is obtained via element-wise alpha blending:
\begin{equation}
  I_{out} = \alpha \odot I + (1 - \alpha) \odot \left( \gamma \cdot (\mathcal{G}_{\sigma^2} \cdot I) \right),
  \label{eq:alpha_blending}
\end{equation}
where $I \in \mathbb{R}^{H \times W}$ represents the original high-fidelity input image, $\alpha \in [0, 1]^{H \times W}$ denotes the continuous soft mask delineating the primary content, $\mathcal{G}_{\sigma^2} \cdot I$ applies a Gaussian smoothing convolution with variance $\sigma^2$ to the background, and $\gamma \in [0, 1]$ acts as a scalar attenuation factor to suppress non-essential structural noise.

\begin{figure}[ht]
  \centering
  \includegraphics[width=\linewidth]{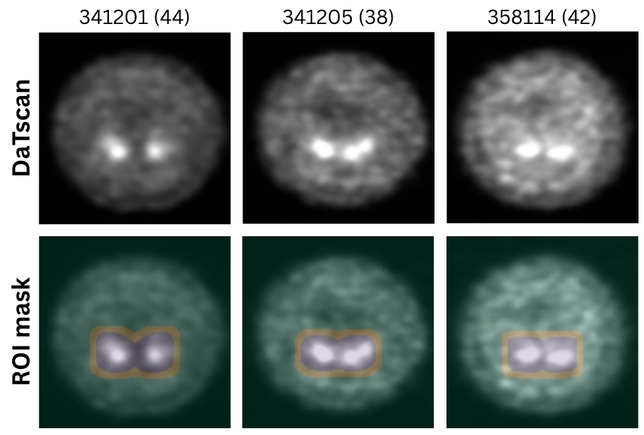}
  \caption{Examples of computed ROI masks. The number above each DaTscan is the image ID, while the number in parentheses is the slice number.}
  \label{fig:roi_mask}
\end{figure}

During the diffusion process, it is computationally inefficient to allocate model resources equally across all 16,384 pixels at every timestep. Our multi-weight region of interest (ROI), shown in \cref{fig:roi_mask}, focuses on the striatum (blue), which occupies approximately 10\% of DaTscan images and is key to understanding PD progression. To capture subtle variations, we added a buffer zone (orange) that accounts for anatomical transitions and potential signal spillover. The remaining area (green) represents background and irrelevant details.

To compute the ROI mask for each patient, we used a spatial agreement approach. A binary mask, $M_i$, is calculated for each screening DaTscan image via adaptive thresholding and morphological filtering of both the original and horizontally flipped views to enforce bilateral symmetry. The striatum is then defined by identifying pixels that satisfy the threshold condition:
\begin{equation}
  \sum_{i=1}^{n} M_i(x, y) \geq n \cdot 0.65,
  \label{eq:roi_calculation}
\end{equation}
where $n$ is the total number of masks for a patient.

By applying a morphological dilation with a 3$\times$3 unit kernel for 4 iterations, we expanded the boundaries of the striatum, creating a buffer zone. In addition to its medical relevance, this ensures numerical stability of diffusion at the edges of the striatum.
\section{Method}
\label{sec:method}

\begin{figure*}
  \centering
  \includegraphics[width=\linewidth]{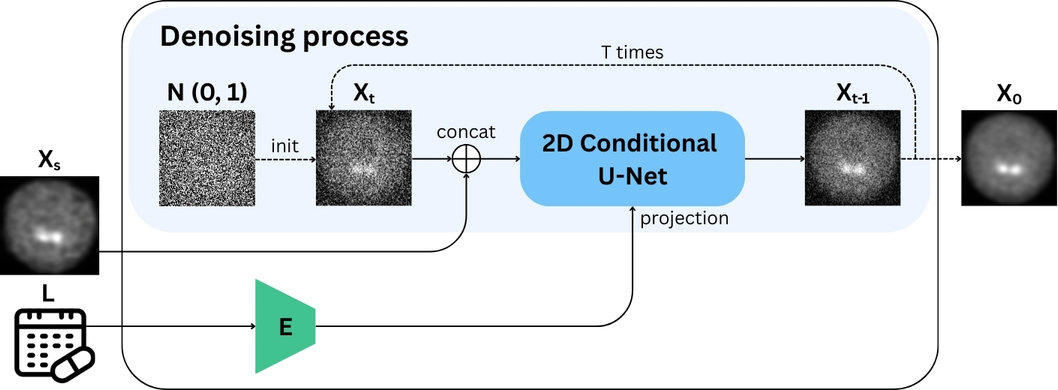}
  \caption{Overview of the treatment-conditioned reverse diffusion process. Starting from Gaussian noise $\mathcal{N}(0, 1)$, the 2D conditional U-Net iteratively denoises the latent representation $X_t$ to forecast the future DaTscan image $X_0$. To preserve original anatomical structure, the screening DaTscan image $X_s$ is channel-wise concatenated with $X_t$ at each timestep. Concurrently, the longitudinal LEDD treatment $L$ is processed by an encoder $E$, and the resulting embedding is projected into the U-Net as a global condition to guide the progression trajectory.}
  \label{fig:diffusion}
\end{figure*}

To forecast the trajectory of PD, we frame the progression modeling as a conditional generation task. The proposed treatment-conditioned diffusion framework learns the structural evolution of the striatum driven by both the original physiological state and the longitudinal pharmacological interventions (see \cref{fig:diffusion}).

Let $X_{s} \in \mathbb{R}^{H \times W}$ denote the screening DaTscan image of a patient, capturing the initial state of the dopaminergic system. Over the subsequent 12-month period, the patient receives a treatment represented by a longitudinal LEDD vector, $L \in \mathbb{R}^{12}$, where each element corresponds to the average daily dose for a given month. Our goal is to predict the future DaTscan image, denoted as $X_0 \in \mathbb{R}^{H \times W}$, exactly one year after the screening. From a probabilistic perspective, we aim to learn the conditional distribution $p_\theta(X_{0} | X_{s}, L)$, allowing to sample future brain states that reflect the natural disease progression. 

The forward diffusion process gradually corrupts the target future DaTscan image $X_0$ by adding Gaussian noise over $T$ timesteps, producing a sequence of noisy latents $X_t$, such that $X_T \sim \mathcal{N}(0, 1)$. To reverse this process and generate the predicted future DaTscan image, we use a conditional 2D U-Net parameterized by $\theta$.

Given that both the screening $X_s$ and target $X_0$ DaTscan images are spatially aligned in MNI152 space during preprocessing, explicit spatial conditioning is applied. At each denoising step $t$, the noisy latent $X_t$ and the screening DaTscan image $X_s$ are concatenated channel-wise to construct the U-Net input. This ensures that the model preserves the persistent anatomical structure of the striatum while modifying only regions affected by physiological changes during PD progression.

Instead of using the traditional noise prediction approach ($\epsilon$-parameterization), our model adopts $v$-parameterization~\cite{salimans2022}. This choice is crucial for training stability and sample fidelity when utilizing few-step sampling or noise schedules that reach a zero terminal signal-to-noise ratio, especially for contrast-heavy medical images~\cite{lin2024}. The velocity target $v_t$ is defined as:
\begin{equation}
  v_t = \alpha_t \epsilon - \sigma_t x_0,
  \label{eq:velocity_target}
\end{equation}
where $\alpha_t$ and $\sigma_t$ are the noise schedule parameters, $x_0$ represents the original sample, and $\epsilon \sim \mathcal{N}(0, 1)$ is the standard Gaussian noise.

\begin{figure}[t]
  \centering
  \includegraphics[width=\linewidth]{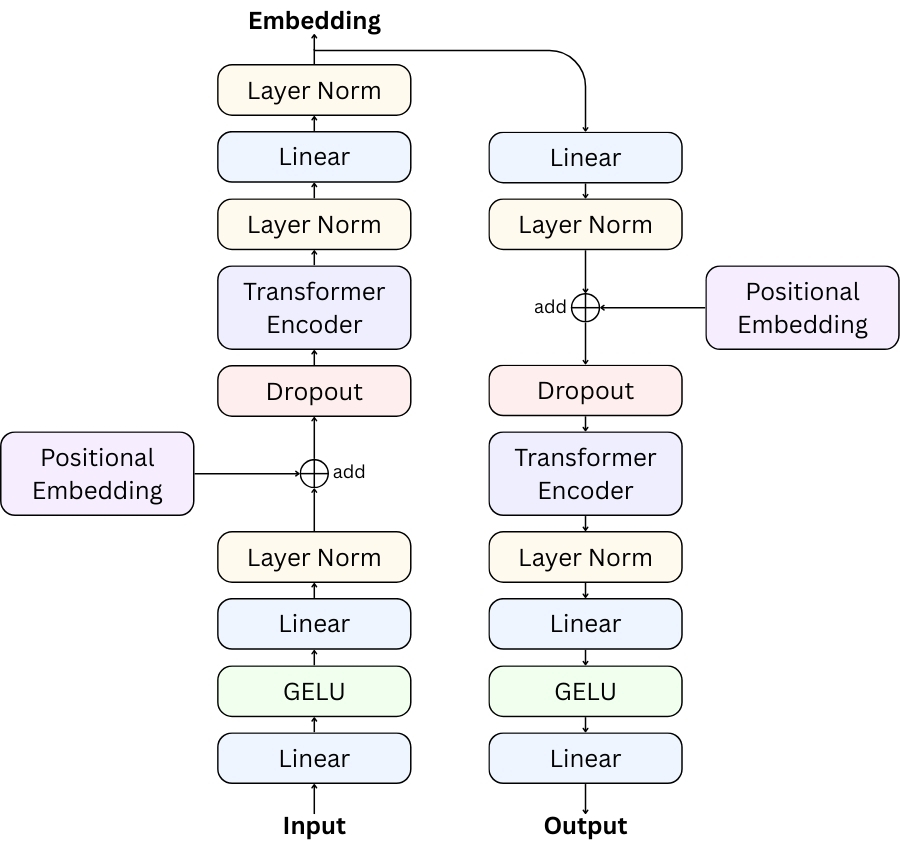}
  \caption{Architecture of the LEDD autoencoder. Continuous longitudinal medication data are projected into a higher-dimensional latent space via linear layers and GELU activations. To capture temporal dynamics, positional embeddings are injected, preserving the chronological order of dosages over the 12-month period. The sequence is then processed by Transformer encoder blocks to represent complex long-range interactions, yielding a unified embedding that is used to condition the diffusion model.}
  \label{fig:autoencoder}
\end{figure}

To condition the velocity prediction on the treatment, the LEDD vector $L$ is first mapped to a dense representation using the encoder of a dedicated Transformer-based autoencoder (see \cref{fig:autoencoder}). Within the U-Net architecture, $E(L)$ is fused with the timestep embedding and injected into the residual blocks via adaptive group normalization (AdaGN), where it modulates feature maps by dynamically predicting scale and shift parameters~\cite{dhariwal2021}.

To prevent the model from allocating capacity to the uninformative background, we penalize errors in the biologically critical regions by applying a weighted mean absolute error (MAE) loss. This step uses a precomputed multi-weight ROI mask $M \in [0, 1]^{H \times W}$, where the coefficient is set to 1.0 for the striatum, 0.8 for the buffer zone around it, and 0.4 for the remaining area.

Modeling pharmacological interventions is challenging due to their time-dependent, non-linear dynamics. To capture these complex temporal patterns, we introduce a Transformer-based sequence-to-sequence autoencoder optimized via a joint reconstruction and contrastive learning objective~\cite{chen2020}. Given that real-world medical records frequently contain missing values and noise, we propose a dual-stage augmentation pipeline to produce semantically meaningful latent representations:

\begin{enumerate}
\item \textit{Clinical block-wise perturbation:} to simulate clinical dose adjustments at the dataset level, we sample time blocks $k \sim \mathcal{U}(3, 6)$ months and apply additive shifts $\Delta = c \times 25$ mg, where $c \sim \mathcal{U}(-5, 5)$.

\item \textit{Continuous stochastic modulation:} during the contrastive training loop, each batch-wise sequence undergoes an additional localized transformation to produce two correlated views, $\tilde{L}^{(1)}$ and $\tilde{L}^{(2)}$, by scaling the sequence by $\gamma \sim \mathcal{U}(0.9, 1.1)$ and injecting Gaussian noise.
\end{enumerate}

The LEDD vectors for both views, $\tilde{L}^{(1)}$ and $\tilde{L}^{(2)}$, are projected into a 256-dimensional hidden space before being processed through a 3-layer pre-norm Transformer block with positional embeddings. A projection head then maps the outputs to their respective latent sequences, $Z^{(1)}, Z^{(2)} \in \mathbb{R}^{12 \times 128}$. During contrastive loss computation, the embeddings are mean-pooled along the temporal axis to yield a single unified representation per view, $r \in \mathbb{R}^{128}$, which is subsequently $L_2$-normalized.

To ensure alignment between the augmented representations, we apply the temperature-scaled information noise-contrastive estimation (InfoNCE) loss ($\mathcal{L}_{CL}$)~\cite{oord2019}. This objective maximizes the pairwise cosine similarity of positive views while repelling negative instances within the batch, utilizing a temperature parameter $\tau = 0.2$. Simultaneously, to ensure the model preserves the exact chronological semantics of the treatment, the decoder reconstructs the original, uncorrupted LEDD vector from the latent space by minimizing the mean squared error (MSE) loss ($\mathcal{L}_{rec}$).

The entire autoencoder is optimized end-to-end via a joint multi-task loss:
\begin{equation}
\mathcal{L}_{total} = \mathcal{L}_{rec} + \beta \mathcal{L}_{CL},
\label{eq:autoencoder_loss}
\end{equation}
where $\beta = 0.1$ acts as a scaling factor to balance the contrastive penalty with the reconstruction fidelity.
\section{Experiments}
\label{sec:experiments}

We implemented the proposed framework in PyTorch and trained it on a single NVIDIA A100 80GB GPU. The diffusion process was configured with a squared-cosine noise schedule~\cite{nichol2021} and a zero-terminal signal-to-noise ratio, utilizing 1000 timesteps for training and 250 denoising steps to forecast DaTscan images during final evaluation. Reflecting our efficiency-oriented setup, the generative backbone is implemented as a lightweight, purely convolutional 2D U-Net. To minimize computational overhead, we avoid expensive cross-attention mechanisms. Instead, the network comprises four downsampling and upsampling levels with feature map dimensions of 128, 256, 512, and 512.

To prevent overfitting on a limited cohort and improve generalization across diverse, multi-center scanner calibrations, we implemented an on-the-fly data augmentation pipeline. It relies on random gamma correction with $\gamma \in [0.8, 1.2]$ to simulate variations in radiotracer uptake and scanner intensity, alongside spatial transformations including random rotations ($\pm 4^\circ$), translations ($\pm 4$ pixels), and scaling ($\pm 5\%$). Crucially, to avoid introducing sharp artificial borders, out-of-bounds regions were padded with the 2nd percentile intensity value rather than absolute zeros. Following these modifications, all DaTscan images were normalized to the $[-1, 1]$ range. Prior to encoding, LEDD vectors were uniformly scaled by $\pm 20\%$ in magnitude to enable continuous interpolation within the treatment latent space, and then normalized via a logarithmic transformation, $L' = \log(1 + L)$, to stabilize high dosage variance and mitigate extreme outliers.

The model was trained for 100 epochs with a batch size of 16 using automatic mixed precision (AMP)~\cite{micikevicius2018}. For optimization, we used the AdamW optimizer~\cite{loshchilov2019} with a weight decay of $10^{-2}$ and a learning rate of $10^{-4}$, which followed a cosine annealing schedule after a linear warmup phase covering the first 10\% of the total training steps. Furthermore, to stabilize the highly non-linear generative dynamics and mitigate color-shift artifacts, we applied an exponential moving average (EMA)~\cite{moralesbrotons2024} with a decay rate of 0.999 to the network weights.

To robustly measure the model's performance, the forecasted DaTscan images were evaluated using three ROI-weighted metrics: structural similarity index measure (SSIM), mean absolute error (MAE), and mean squared error (MSE). As a baseline, we used the no-progression assumption, which treats the screening DaTscan image as identical to the month 12 DaTscan image.

\begin{figure*}
  \centering
  \includegraphics[width=\linewidth]{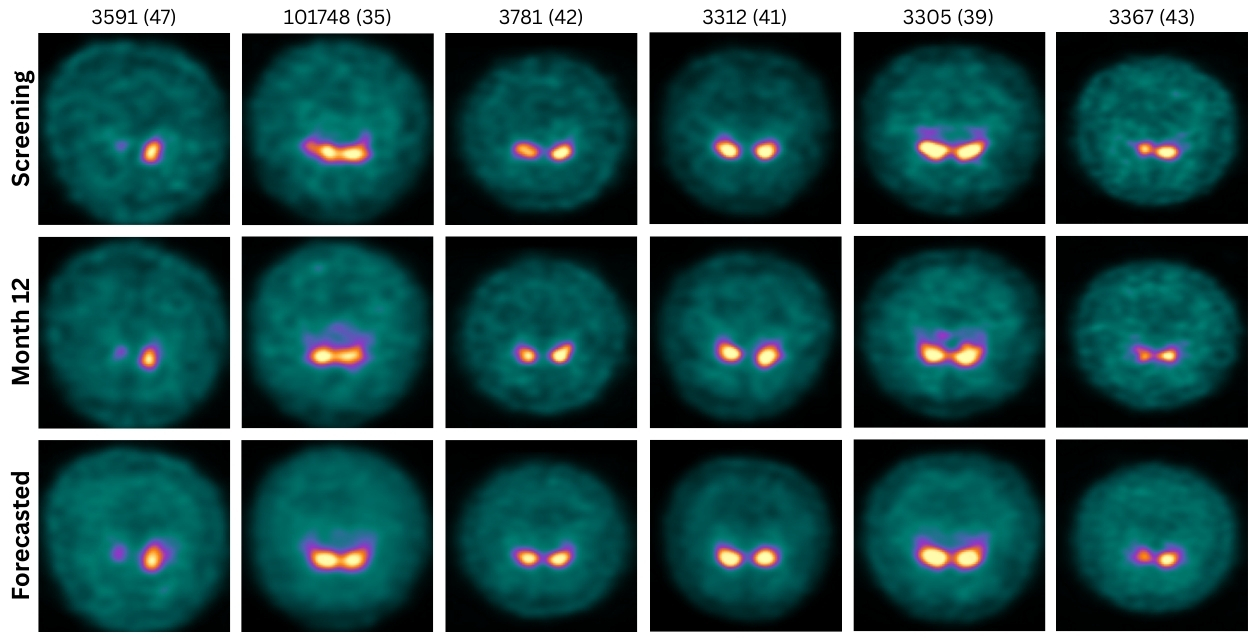}
  \caption{Qualitative comparison of Parkinson’s disease (PD) progression forecasting. The figure demonstrates the visual fidelity of the treatment-conditioned diffusion model across different subjects. Screening: baseline DaTscan images used as the spatial condition. Month 12: ground-truth DaTscan images acquired at month 12. Forecasted: future DaTscan images predicted using the model. The number above each column is the subject ID, while the number in parentheses is the slice number.}
  \label{fig:forecasting}
\end{figure*}

For visual verification of generative fidelity and clinical plausibility, we present a side-by-side comparison of the screening, ground-truth month 12, and forecasted month 12 DaTscan images, rendered using pseudocolor mapping (see \cref{fig:forecasting}). Unlike traditional models that suffer from regression-to-the-mean blurring~\cite{rassmann2025}, our $v$-parameterized approach predicts artifact-free future brain states while preserving sharp structural boundaries.

PD typically presents with clearly prominent asymmetric dopaminergic dysfunction, shown by a significantly greater reduction in radiotracer uptake in one cerebral hemisphere compared to the other. The forecasted DaTscan images accurately reflect this unilateral degradation (\eg, in subject 3591) rather than applying a naive, uniform global dimming effect.

The introduced model consistently outperforms the static baseline across all depth variations of the basal ganglia (see Table \ref{tab:model_metrics}). On average, our method achieves a robust $+4.9\%$ relative improvement in structural fidelity (increasing SSIM from $0.658$ to $0.690$) while reducing MAE and MSE by $7.2\%$ and $14.0\%$, respectively. This improvement indicates that overall anatomical shape is preserved while localized changes are accurately captured.

Notably, the most substantial MSE improvements occur at the anatomical boundaries of the striatum (\eg, $18.4\%$ for slice 34 and $17.6\%$ for slice 43), where morphological decay and shape contraction are often most pronounced. The consistent superiority across all slices confirms that the model successfully learns the 3D spatial coherence of the disease progression, despite operating on 2D cross-sections. Conversely, the central slices (\eg, 39 to 41), which encompass the bulk of the striatal mass and exhibit less pronounced shape variation over the course of a year, show a more moderate yet consistent improvement.

\begin{table*}
  \caption{Quantitative evaluation of the forecasted DaTscan images over a 12-month period. Performance is measured using ROI-weighted structural similarity index (SSIM), mean absolute error (MAE), and mean squared error (MSE) across axial slices 34 to 47. The baseline represents the static assumption of no morphological progress from the screening DaTscan image. The $\Delta$ columns indicate the achieved relative percentage improvement.}
  \label{tab:model_metrics}
  \centering
  \renewcommand{\arraystretch}{1.2} 
  \setlength{\tabcolsep}{6pt}
  \begin{tabular}{cccccccccc}
    \toprule
    \multirow{2}{*}[-0.25em]{Slice} & \multicolumn{3}{c}{\textbf{Baseline}} & \multicolumn{6}{c}{\textbf{U-Net}} \\
    \cmidrule(lr){2-4} \cmidrule(lr){5-10}
    & {SSIM $\uparrow$} & {MAE $\downarrow$} & {MSE $\downarrow$} 
    & {SSIM $\uparrow$} & {$\Delta$ SSIM} 
    & {MAE $\downarrow$} & {$\Delta$ MAE} 
    & {MSE $\downarrow$} & {$\Delta$ MSE} \\
    \midrule
    $34$ & $0.632$ & $0.101$ & $0.021$ & $0.669$ & $+5.9\%$ & $0.091$ & $-9.2\%$ & $0.017$ & $-18.4\%$ \\
    $35$ & $0.641$ & $0.102$ & $0.021$ & $0.678$ & $+5.8\%$ & $0.093$ & $-8.5\%$ & $0.018$ & $-15.8\%$ \\
    $36$ & $0.650$ & $0.101$ & $0.022$ & $0.684$ & $+5.2\%$ & $0.094$ & $-7.6\%$ & $0.018$ & $-15.7\%$ \\
    $37$ & $0.661$ & $0.099$ & $0.021$ & $0.691$ & $+4.5\%$ & $0.093$ & $-6.1\%$ & $0.018$ & $-13.2\%$ \\
    $38$ & $0.669$ & $0.097$ & $0.020$ & $0.697$ & $+4.2\%$ & $0.092$ & $-5.4\%$ & $0.018$ & $-12.5\%$ \\
    $39$ & $0.673$ & $0.096$ & $0.020$ & $0.698$ & $+3.8\%$ & $0.092$ & $-4.2\%$ & $0.018$ & $-9.5\%$ \\
    $40$ & $0.672$ & $0.096$ & $0.019$ & $0.696$ & $+3.6\%$ & $0.091$ & $-4.2\%$ & $0.018$ & $-8.8\%$ \\
    $41$ & $0.669$ & $0.095$ & $0.020$ & $0.696$ & $+3.9\%$ & $0.090$ & $-5.6\%$ & $0.018$ & $-10.3\%$ \\
    $42$ & $0.665$ & $0.095$ & $0.019$ & $0.690$ & $+3.7\%$ & $0.090$ & $-5.5\%$ & $0.018$ & $-9.7\%$ \\
    $43$ & $0.660$ & $0.095$ & $0.019$ & $0.695$ & $+5.3\%$ & $0.086$ & $-9.6\%$ & $0.016$ & $-17.6\%$ \\
    $44$ & $0.657$ & $0.094$ & $0.019$ & $0.695$ & $+5.8\%$ & $0.085$ & $-9.7\%$ & $0.016$ & $-17.9\%$ \\
    $45$ & $0.655$ & $0.094$ & $0.019$ & $0.691$ & $+5.5\%$ & $0.086$ & $-8.8\%$ & $0.016$ & $-16.2\%$ \\
    $46$ & $0.654$ & $0.092$ & $0.018$ & $0.691$ & $+5.8\%$ & $0.085$ & $-7.9\%$ & $0.016$ & $-14.2\%$ \\
    $47$ & $0.654$ & $0.090$ & $0.018$ & $0.694$ & $+6.1\%$ & $0.082$ & $-9.1\%$ & $0.015$ & $-15.9\%$ \\
    \midrule
    \textbf{Mean} & $\mathbf{0.658}$ & $\mathbf{0.096}$ & $\mathbf{0.020}$ & $\mathbf{0.690}$ & $\mathbf{+4.9\%}$ & $\mathbf{0.089}$ & $\mathbf{-7.2\%}$ & $\mathbf{0.017}$ & $\mathbf{-14.0\%}$ \\
    \bottomrule
  \end{tabular}
\end{table*}

Despite the significant improvements achieved by our framework, forecasting the exact morphological evolution of DaTscan images remains a highly complex and non-trivial task. While LEDD provides a crucial, measurable metric of pharmacological intervention, the underlying progression of PD is multifactorial and profoundly heterogeneous~\cite{Abusrair2022}. The uptake of radiotracers in the striatum is not strictly a deterministic function of the screening anatomical state or medication history. Instead, it is heavily influenced by a broad spectrum of unobserved biological, genetic, and environmental variables.

A major limitation is that patients differ in their genetic profiles and clinical presentations. Certain genetic mutations can affect the rate of dopaminergic neurodegeneration~\cite{Aasly2020}. In addition, patients exhibit different clinical forms of the disease, such as tremor-dominant or akinetic-rigid types, which are associated with distinct rates of brain structural changes. Because the model lacks genetic information or detailed clinical subtypes, these unmeasured factors contribute extra variability in basal ganglia changes, which cannot be fully explained by overall medication alone.

Additionally, LEDD serves as a macroscopic proxy for treatment, but it fails to capture the detailed pharmacokinetics of levodopa metabolism. Personal differences in gastrointestinal absorption, blood-brain barrier permeability, and even dietary habits (\eg, high protein intake competing with levodopa absorption) significantly affect the actual efficacy of the prescribed dose~\cite{Leta2023,Rusch2023}. Therefore, in clinical practice, two patients with identical screening DaTscan images and LEDD treatments may exhibit different physiologic responses, leading to distinct follow-up DaTscan images.
\section{Conclusion}
\label{sec:conclusion}

In this work, we introduced a novel treatment-conditioned diffusion framework for forecasting the progression of PD. By shifting the predictive paradigm from scalar clinical scores to high-fidelity synthesis of future brain states, our approach addresses the loss of anatomical details and the blurring common in earlier methods. We demonstrated that conditioning a 2D U-Net on screening DaTscan images and LEDD treatments, processed through a Transformer-based encoder, successfully captures the complex, non-linear dynamics of pharmacological interventions and natural morphological decay.

Our results show the clinical and quantitative strength of this method. By using a multi-weight ROI optimization pipeline, the framework efficiently allocates its representational ability to biologically critical regions, such as the striatum. Consequently, the model significantly outperforms the no-progression baseline, achieving a 14.0\% reduction in MSE, a 7.2\% reduction in MAE, and a 4.9\% improvement in SSIM across 14 slices. Crucially, qualitative analyses confirm that our pipeline prevents regression-to-the-mean blurring and accurately captures highly localized, asymmetric dopaminergic dysfunction while maintaining sharp structural boundaries.

Moving forward, this generative approach offers a solid foundation for customized therapeutic planning, potentially allowing clinicians to simulate counterfactual treatment trajectories to improve outcomes. By providing a reliable way to visualize patient-specific morphological decay before severe symptomatic onset, this framework empowers medical professionals to create more proactive and targeted clinical interventions.

The proposed model unifies the gap between state-of-the-art conditional diffusion modeling and clinical comprehensibility in neurodegenerative disease forecasting, affording a key step towards data-driven medicine in neurology. Future work would focus on extending the predictive horizon, scaling the architecture to handle 3D neuroimaging modalities, and grounding the generative process in established biological priors.
{
    \small
    \bibliographystyle{ieeenat_fullname}
    \bibliography{main}
}

\end{document}